# AI Coding with Few-Shot Prompting for Thematic Analysis

*Samuel Flanders[1], Melati Nungsari[1], Mark Cheong Wing Loong[2]*

[1] *Asia School of Business, Kuala Lumpur, Malaysia*
[2] *Malaysia School of Pharmacy, Monash University*

**Abstract**

This paper explores the use of large language models (LLMs), here represented by GPT 3.5-Turbo to perform coding for a thematic analysis. Coding is highly labor intensive, making it infeasible for most researchers to conduct exhaustive thematic analyses of large corpora. We utilize few-shot prompting with higher quality codes generated on semantically similar passages to enhance the quality of the codes while utilizing a cheap, more easily scalable model.

# Introduction

This paper explores the use of large language models (LLMs), here represented by GPT 3.5-Turbo (henceforth "GPT"), to perform coding for a thematic analysis. Coding is highly labor intensive, making it infeasible for most researchers to conduct exhaustive thematic analyses of large corpora.

Recent advances in large language models (LLMs) have opened the door to novel approaches for automating aspects of qualitative research, including thematic analysis (TA). Prior work has shown that LLMs can generate plausible thematic codes for text data (Dai, Xiong, and Ku, 2023; Morgan, 2023; De Paoli, 2024). This paper focuses on the development and evaluation of an AI-assisted coding methodology designed to enhance the thematic coding of text passages using large language models. Our goal is to improve the quality of machine-generated codes through careful Chain-of-Thought (CoT) prompt engineering and few-shot learning.

Our test case is a corpus of 2,530 Malaysian news articles on refugees, collected to explore the research question: *What attitudes towards refugees are present in Malaysian news media?* Each article was segmented into passages and coded by GPT using a structured, multi-step process. Early iterations revealed a range of challenges—such as irrelevant or ambiguous passages, "summary bleeding" where the model's response relied on information outside the target text, and overgeneralized or misattributed codes. To address these, we refined the prompts, implemented exclusion criteria, and introduced a **Socratic prompting** framework that walks the model through relevance checks and structured CoT reasoning.

Crucially, we also introduce a scalable form of few-shot prompting. Rather than providing random examples to guide GPT's output, we cluster semantically similar passages using vector embeddings and provide GPT with coding summaries from exemplar passages in the same cluster. These exemplars are generated through our longform Socratic pipeline in our study, but could be supplied by human researchers. This approach leverages the benefits of few-shot

prompting—improved consistency and specificity—while remaining computationally efficient and theoretically grounded.

We evaluate the performance of our methodology by asking three human coders to independently assess a fixed sample of coded passages across two iterations: an initial version and a final version incorporating the improvements described above. Reviewers rated the appropriateness of the codes, identified irrelevant passages, and engaged in structured discussions to resolve disagreements. The final version achieved a high level of consensus, with an F1 score above 0.82 and a negative predictive value of 0.97—suggesting that the method is especially strong at identifying irrelevant material. Inter-rater reliability also improved substantially between iterations.

Our contribution in this paper is twofold: first, we provide a detailed, replicable methodology for improving the quality of LLM-generated thematic codes using prompt refinement, error analysis, and structured few-shot prompting; second, we evaluate this methodology through systematic human review, identifying both its strengths and limitations. This work lays the foundation for future applications of AI coding in qualitative research that preserve the interpretive richness of human analysis while benefiting from the scalability and consistency of machine assistance.

# Literature Review

The integration of AI into qualitative research has prompted a rethinking of how thematic analysis is conducted. Traditional manual coding remains a cornerstone of qualitative research, with methods ranging from organic, interpretive procedures to more structured, positivist approaches. Early work in qualitative software—such as NVivo (Dawborn-Gundlach & Pesina, 2015)—has complemented manual coding, yet these methods typically demand intensive researcher involvement to interpret meaning and develop themes.

### *Coding and Thematic Analysis Approaches*

Variations in thematic analysis reflect differing assumptions about the role of the researcher and the nature of data interpretation. On one end of the spectrum, approaches like Coding Reliability (Boyatzis, 1998) treat codes as succinct domain summaries, effectively converting qualitative data into quantifiable measures. Such methods align naturally with a positivist perspective, in which replicability and objective measurement are paramount. In contrast, flexible methods—exemplified by Clarke and Braun (2006)—stress an organic evolution of themes where the researcher's interpretive input is critical. Furthermore, critical realist perspectives (Fryer, 2022; Christodoulou, 2023) introduce frameworks that combine abductive reasoning and causal explanation, thereby expanding the repertoire of interpretive strategies. While these interpretivist methods have enriched qualitative inquiry, they also introduce subjectivity that may reduce reproducibility across studies.

### *AI and Thematic Analysis*

Recent advances in large language models (LLMs) have enabled a new strand of research in thematic analysis. Studies by Dai, Xiong, and Ku (2023) and Morgan (2023) have demonstrated that LLMs are capable of producing descriptive codes that capture concrete aspects of textual data. Other methodologies integrate LLM outputs with human oversight—through iterative discussions and code refinement—to approximate the reliability of dual human coders (Chew et al., 2023; De Paoli, 2024; Turobov et al., 2024; De Paoli and Mathis, 2024). In contrast to these models, the approach presented in this paper leverages few-shot prompting to generate codes that are explicitly data-derived and reproducible. Rather than emphasizing the interplay between human interpretation and automated suggestions, our method focuses on producing objective codes that are both measurable and scalable, thereby aligning with the positivist notion of generating quantifiable evidence.

### *Synthesis and Implications for AI Coding Methodologies*

The literature reveals a tension between interpretive flexibility and objective rigor in thematic analysis. While traditional methods—whether reflexive or critical realist—stress the indispensability of human insight (Clarke & Braun, 2017; Fryer, 2022; Christodoulou, 2023), positivist approaches such as Coding Reliability (Boyatzis, 1998) underscore the benefits of reproducible, data-derived codes. In this context, our methodology—centered on few-shot prompting and a structured, Socratic coding process—represents a hybridization of these approaches. By focusing on generating concrete and measurable codes, the method circumvents many subjective pitfalls associated with saturation and interpretive variability. It thereby provides an avenue for scaling qualitative analysis, enabling researchers to assess large datasets with a degree of consistency that supports subsequent statistical validation and theoretical development.

# Methodology

### *Data collection*

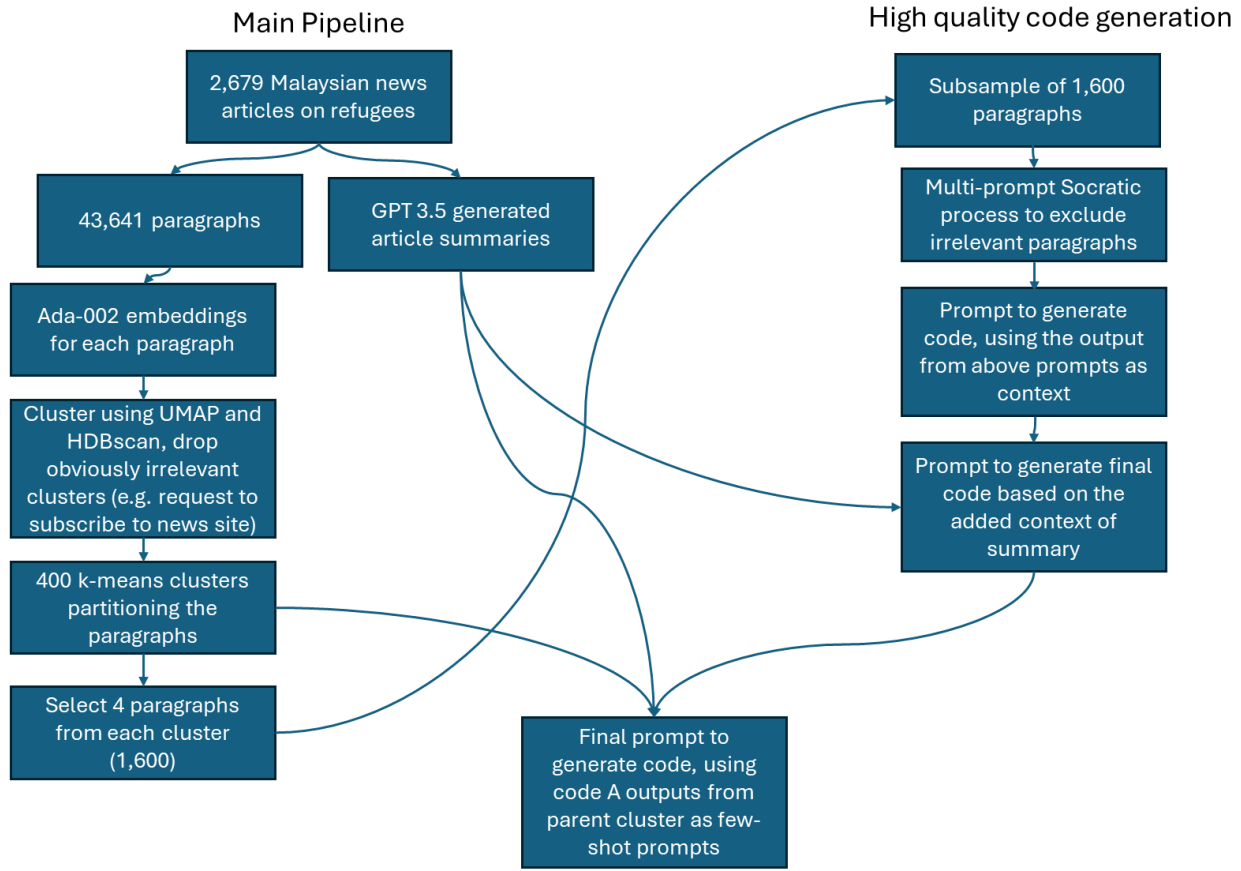

*Figure 1: Data Collection and Coding Process Diagram*

Our thematic analysis focused on attitudes towards refugees in Malaysian media, so we chose to build our corpus from Malaysian news articles. To choose sources, we performed a google search for "malaysia news" and excluded promoted results, with a target of five distinct sources to ensure variety. Many search results were unusable for various reasons shown in Table 1, so our five sources come from the top 11 results.

| Source | Included | Notes |
| --- | --- | --- |
| The Star | Yes | Data only available from Jan 2020 |
| Malay Mail | Yes | |
| Malaysiakini | Yes | |
| FMT | No | Scraping not allowed |
| NST | No | Scraping not allowed |
| MalaysiaNow | No | Search query only allows "or" logical operator, not "and" so malaysia refugees generates largely irrelevant results. |
| Yahoo Malaysia | No | Just an index of other news sources |
| Bernama | No | Search results only for past 2 weeks |
| TheMalaysianInsight | Yes | |
| Al Jazeera | No | Not local |
| The Sun | Yes | |

*Table 1: News Sources*

To collect the data, we performed the query '"malaysia" "refugees"' for each source's website search and collected all articles from January 1st, 2017 to the date of search, which varied from March to May 2023. To generate the passages of our corpus, we split the articles by line break and dropped all passages less than 50 characters long. We also generated one paragraph summaries of each article using GPT in order to provide context to GPT when it analyzed the passages.[1]

### *AI Coding I*

In our first iteration, for each passage we prompted GPT with the following:

```
'Read a passage from a news article summarized here: ### ' + str(summary) + '
### passage: ### ' + str(excerpt)+ ' ### In 12 words or less, give the theme of
this specific passage as it embodies, relates to or reflects attitudes towards
refugees in Malaysia, or return "Irrelevant"'
```

Where "summary" contains the one paragraph summary of the full article and "excerpt" contains the passage. An example will be instructive:[2]

---

[1] Many passages are ambiguous without context, e.g. "They should be sent home" is relevant to attitudes towards refugees only if "They" refers to refugees. However, each passage is analyzed with a single prompt and the cost of using the OpenAI API is proportional to the number of words in a prompt, so including the full article as context would be prohibitively expensive. A short summary was chosen to provide context with far fewer words.

[2] Ten more examples, randomly selected from the corpus, can be found in Table A1.

**Summary:** *The article discusses Malaysia's treatment of refugees and asylum seekers, contrasting it with the country's past efforts to assimilate Bosnian refugees in the 1990s. Malaysia is not a signatory of the 1951 Refugee Convention, which means that refugees and economic immigrants without legal documentation are treated the same under existing laws. This leaves refugees exposed to harsh penalties, mistreatment, exploitation, and discrimination. The article argues for Malaysia to begin the process of allowing refugees to work legally and to be given the opportunity to be educated in the country. The article also highlights the economic benefits of granting refugees the right to work and the lack of access to education for refugee children.*

**Excerpt:** *The same decade saw Malaysia, collectively with many other nations actively assimilating Bosnian refugees into our communities, providing them with protection and educational opportunities they desperately needed.*

**Coding:** *Positive past treatment of refugees in Malaysia contrasted with current situation.*

Further, we embedded these codings (Coding) using OpenAI's *Ada* embedding. This generated a 1536 element vector capturing the *semantic meaning*[3] of the coding, which we projected down to 6 dimensions using UMAP and clustered using HDBSCAN. The end result was 60 clusters of passages, where each cluster was comprised of passages whose codes were semantically similar.

## *Coding Review*

We then randomly sampled 414 passages from the corpus and asked three reviewers to assess the coding on a Likert scale of (Strongly Disagree, Disagree, Unsure, Agree, Strongly Agree), as well as provide comments when relevant. Coding documents are available at URL. The coding document provided reviewers with the full text of the article, a link to the article URL, the title, the same summary ChatGPT was provided, the excerpt itself, the GPT generated code, and the cluster of the passage.[4]

Reviewers were provided with the following instructions:

*You will be provided with a Google Sheet that contains a sampling of passages from Malaysian news articles on refugees from 2017-2023. This document also contains summaries of the articles, titles, and the full text of each article.*

---

[3] The 1536 dimensions of the vector represent different facets of a passage's meaning, as understood by the Ada LLM. Two passages with similar meaning will be mapped to points close together in the 1536-dimensional space, while passages that differ in meaning will be far apart. This is of course limited by the capacity of Ada to assess meaning.

[4] Cluster index was included to ease assessment—the sample was sorted by cluster and it was hoped that grouping passages by similar codes would make assessment easier.

*ChatGPT was asked to code each passage "as it embodies, relates to or reflects attitudes towards refugees in Malaysia". Your task is to assess the appropriateness of the codes ChatGPT generated for each passage.*

*Specifically, for each passage, to what extent to [sic] you agree with ChatGPT's coding? (1. Strongly Disagree 2. Disagree, 3. Unsure, 4. Agree, 5. Strongly Agree)*

*It is up to you to choose a value from 1 to 5. There is no formal criteria for e.g. "strongly agree". If you're not sure that you strongly agree, rather than worrying about exactly which value (1-5) is most appropriate, provide comments on what aspects of the coding you disagree with and provide an example of how you would have coded it.*

*Complete this assessment independently and do not discuss your findings until all reviewers have completed this task.*

Reviewers were also provided significant clarifications throughout the process. These communications are available in appendix N.

### *Methodology Iteration*

As this project was intended to develop a novel method of AI coding, we planned to revise our approach after the first review. We convened a discussion with the reviewers to elicit their feedback on how well ChatGPT coded the given passages. We gave special focus to instances where the reviewers were unanimous in their disagreement with ChatGPT's coding, as well as instances where the reviewers struggled to assess the appropriateness of the coding. Using these instances, we asked the reviewers to identify issues, concerns, and problems with the results of ChatGPT's coding. The reviewers identified several issues from their individual perspectives and we were also able to synthesize additional issues from the discussions between the reviewers. Key issues included the following:

1. Summary bleeding: the code often reflected content from the summary not present in the passage to be coded.

2. Short passages did not code well.

3. General inaccuracy.

4. The term "attitudes" was used very broadly in the codes—often covering cases that weren't attitudes or assigning the attitude to the wrong person.

5. Two identical pronouns (two he's or two she's for two different individuals) in the passage led to confusion.

6. GPT often missed sarcasm in the passage.

7. In passages with a complex logical structure, e.g., describing a relationship between several different individuals or concepts, GPT often made errors in interpretation.

Additionally, they identified common types of passages in the corpus that frequently miscoded. Specifically, these types of passages were never relevant to attitudes towards refugees in Malaysian media but were often erroneously assigned codes. These included:

1. Paragraphs introducing an opinion in an op-ed.

2. Descriptions of organizations.

3. Requests to subscribe to the news site.

4. Short, incomplete sentences.

5. Photo captions.

6. Descriptions of the next article.

For these passage types, our goal was to simply identify them and exclude them from the corpus as they were never relevant. Using the clustering technique with UMAP and HDBscan described above we created an even more granular partition of the corpus into clusters, identified the clusters corresponding to the above passage types, and dropped them from the corpus.

While removing irrelevant passage types was easy, many of the general issues described before could not be directly addressed: issues 5, 6, and 7 are essentially limitations of GPT and the basic methodology. However, we were able to make many improvements. To address 1, 2, and 3 we increased the minimum passage length from 50 to 100 characters—short passages were both the worst coded and least relevant. For 1., we slightly shortened the summaries so they would not overshadow the passage. To address 1. and 4. we tested many alternate prompts and arrived at one that better directed GPT with regard to separating the summary from the passage to be coded and what an attitude is.

Additionally, to address 3. and 4., we made a qualitative change to our prompting approach. Instead of simply asking for a code, we asked for a structured list of 4 responses, seen at the end of the prompts below. GPT is very good at responding in standard programmatic formats, so the JSON format[5] was chosen for the prompt.

Finally, we added an additional element to our methodology. It seemed clear that we could get better coding from a more involved, Socratic approach where the we repeatedly queried GPT on various aspects of the coding process, progressively built up some preliminary assessments of the passage, and then asked GPT to generate the final coding using its earlier, guided assessment as a reference. The reasons were twofold:

---

[5] JSON responses are now a standard option in the OpenAI API, but at the time of analysis these were generated with a request for JSON in the prompt.

1. This allowed us to more carefully and granularly guide the coding process, ensuring GPT considered each facet of the process and potential stumbling block.
2. GPT's capacity to execute multiple tasks in a single response is limited. Breaking tasks down into simpler pieces leads to better performance on each piece.

This approach, of course, requires many times more computing resources per passage. Given contemporaneous API call prices, it was not scalable to the full dataset. To address this, we

1) Partitioned the full dataset into 400 clusters,[6]
2) Sampled four passages from each cluster,
3) Coded using the longform Socratic approach for those 1600 passages, and
4) For each of the 400 clusters, generated a summary of the four codes from the four passages in that cluster.
5) When generating our final codings on the full dataset, identified which of the 400 clusters that passage belonged to and provided the coding summary from 4) as context to GPT.

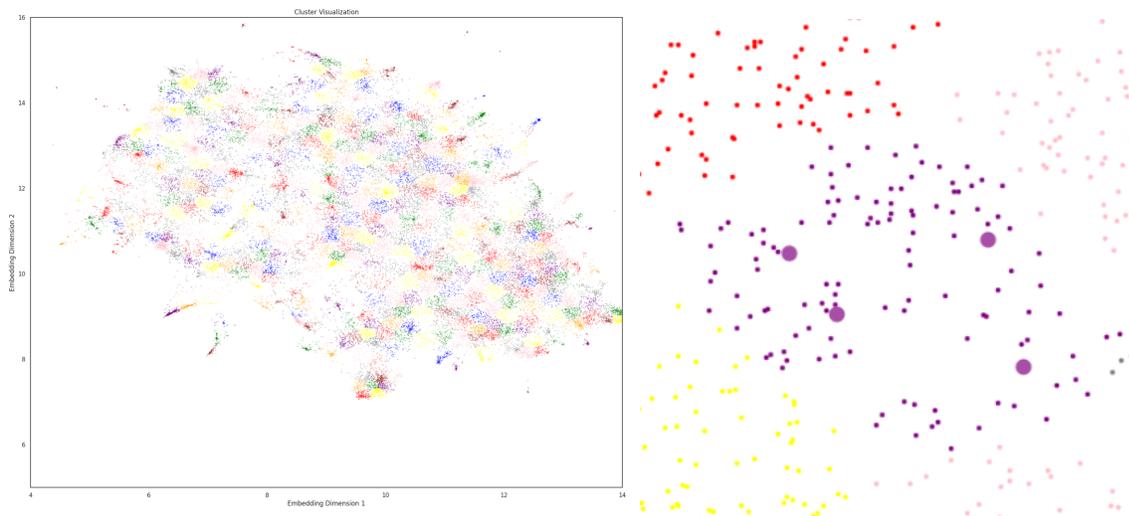

Figure 2, A: 2D Visualization of the 400 clusters. B: Detail of one cluster with the four sampled passages highlighted

The theory for this approach is that passages in the same cluster should be similar in many ways, and seeing high quality codes for similar passages as references improve GPT's performance. This can be thought of as a form of few-shot prompting, where an AI model is provided with examples of the desired output to guide its answer. Due to the high cost of input text, we summarized the four codes to make the prompt more concise. In this paper, we generate these examples with an AI-based process, but practitioners may also consider hand coding a subset of the corpus to use for few-shot prompting.

---

[6] These were distinct from the previous clusters and used the k-means method rather than HDBScan, Our main concern here was creating a complete, relatively symmetric partition, which HDBScan does not offer. HDBScan leaves a large section of the dataset unclustered, only clustering tightly grouped subsets of data while k-means puts every datapoint in a cluster.

The longform Socratic approach was as follows: we asked GPT to assess whether the passages were photo captions or a disclaimers:

```
'Read a passage from a news article ### ' + str(excerpt)+ ' ### Is this passage a piece of text such as 1. a disclaimer of opinion, 2. a photo caption. Or is it a complete passage from the body of a news article? Respond only in the following python dictionary format: {"1. disclaimer?": True/False, "2. caption?": True/False, "Body?": True/False }'
```

Then we asked GPT to assess whether the passages explicitly discuss refugees and Malaysia:

```
'Read a passage from a news article ### ' + str(excerpt)+ ' ### Step by step, answer the following questions: 1. Does the passage explicitly, unambiguously discuss refugees? Note: most passages are not about refugees. 2. Does the passage explicitly, unambiguously reference Malaysia? Note: most passages are about other countries.' + note + ' Now respond in the following Python dictionary format: {"1. Refugees?": "Yes."/"No.", "2. Malaysia?": "Yes."/"No."}'
```

The *note* variable feeds GPT's analysis in the previous step(s) back into the current step. If no red flags were raised, it is empty: "", however, if any red flags were raised, it summarizes the previous issues. In this step it will flag the possibility that "Passage is a disclaimer of personal opinion" or "Passage is a photo caption".

Next, we elicited GPT's confidence that the passage was actually relevant and why not if not:

```
'Read a passage from a news article ### ' + str(excerpt)+ ' ### Answer step by step: 1. Might this passage be relevant to attitudes towards refugees in Malaysia? If it clearly is, answer "Yes." If it might be, depending on the context of the article the passage is from--eg. the identity of the subject and their location--answer "Maybe." If it is definitely irrelevant regardless of context, answer "No." ' + note + ' 2. If "No." or "Maybe.", in 15 words or less give any and all reasons why it might be irrelevant--both those provided earlier and any others you identify, such as irrelevant output from a content management system or editorial annotations to the article. Respond in the following python dictionary format: {"1. Relevant?":"Yes."/"Maybe."/"No.","2. Why Not?":string or None} '
```

Here, the *note* variable is empty if no red flags were raised: "", however, if any red flags were raised, it adds the following text to the prompt: *'Note: this passage has been flagged as possibly meeting the following criteria for irrelevance: [CRITERIA] ### If any of these criteria are true, you should answer "No." or "Maybe. "*, where [CRITERIA] are given as *'Passage is a disclaimer of personal opinion, ', ' Passage is a photo caption, ',' Not about refugees, ', ' Not about Malaysia, '* as relevant.

In the fourth prompt, we ask GPT to code the passage, taking into account its previous assessment of relevance.

```
'Read a passage from a news article ### ' + str(excerpt)+ ' ### Give the theme of this passage as it embodies, relates to or reflects attitudes towards refugees in Malaysia if it is relevant to that topic. If it is not relevant, simply summarize the passage in a few words. Note that this passage may simply be text from the web interface and not from an article at all. Before answering, analyze step by step: 1. in 14 words or less, return the theme. Do not offer a generic theme like "attitudes towards refugees in Malaysia", but give a specific theme. 2. Whose attitudes are being reflected? Examples: the
```

```
Malaysian government, The Bangladeshi government, Malaysians, NGOs, the author. 3. Who is the target of
the attitudes? Examples: migrant workers, Myanmar, the Rohingya, the government, UNHCR. 4. What is the
valence of attitudes towards the target, if any?:  "Sympathetic.", "Hostile.", or "N/A". ' + note + '
Finally, Respond ONLY in the following python dictionary format: {"1. Theme": stringval1, "2. Whose
Attitude?":stringval2,"3. Target":stringval3,"4. Valence": "Sympathetic."/"Hostile."/"N/A"}'
```

In this prompt, the variable *note* is empty if "Relevant?" was assessed as "Yes" in the previous step, *' Previous analysis found that this passage might be irrelevant for this reason: ' + reason + '### Take this into account.'* if "Relevant?" was "Maybe", and *' Previous analysis found that this passage is irrelevant for this reason: ' + reason + '### Take this into account.'* if "No", where *reason* was the reason given in "Why Not?" in the previous step.

Note that, in the Socratic approach, the initial coding does not reference the summary. Instead, we first see if the passage stands on its own as relevant, then ask GPT to reassess its code in light of the summary. This helps prevent summary bleeding, where GPT's coding is only relevant to the summary, not the passage of interest. This leads us to our final coding step:

```
'Read a passage from a news article ### ' + str(excerpt)+ ' ### The theme of this passage was coded as
### ' + str(precode) + ' ### but this analysis ignores the article summary and is therefore unreliable.
Reassess the theme of this passage as it relates to attitudes towards refugees in Malaysia, given the
context of this summary of the article it came from ### ' + str(summary) + ' ### Before answering,
analyze step by step: 1. in 14 words or less, return the reassessed theme (if relevant) as it relates
to attitudes towards refugees in Malaysia, or return None. Do not give a generic theme like "attitudes
towards refugees in Malaysia", but provide a specific theme. If irrelevant, return None for all further
questions. If relevant, 2. Whose attitudes are being reflected? Examples: the government, Malaysians,
NGOs, the author. 3. Who is the target of the attitudes? Examples: the Rohingya, the government, UNHCR.
4. What is the valence of the attitude towards the target, if any?:  "Sympathetic.", "Hostile.", or
"N/A". ' + note + ' Once again, the passage to code is ### ' +  str(excerpt)+ ' ### Finally, Respond
ONLY in the following python dictionary format: {"1. Theme": stringval1/None, "2. Whose
Attitude?":stringval2,"3. Target":stringval3,4. Valence": "Sympathetic."/"Hostile."/"N/A"}'
```

In this prompt, *precode* is the code from the previous step and *note* contains *' Previous analysis found that this SPECIFIC passage might be irrelevant for this reason: ' + reason + '### Does the summary clarify this?.'*

Or *' Previous analysis found that this SPECIFIC passage might be irrelevant for this reason: ' + reason + '### Does the summary clarify this?.'*

If relevance was assessed as maybe or no, respectively.

Here is an example of the four codes for one of the clusters:

{"1. Theme": "Hope for citizenship and repatriation", "2. Whose Attitude?": "The speaker", "3. Target": "Burmese government", "4. Valence": "Sympathetic."}

{"1. Theme": "Praise for repatriation pact and arrangements", "2. Whose Attitude?": "The speaker", "3. Target": "Myanmar and Bangladesh", "4. Valence": "Sympathetic."}

{"1. Theme": "Importance of citizenship for Rohingya refugees in Malaysia", "2. Whose Attitude?": "Saifuddin", "3. Target": "Rohingya refugees", "4. Valence": "Sympathetic."}

{"1. Theme": "Desire for humane treatment and living conditions", "2. Whose Attitude?": "Rohingyas", "3. Target": "Return home with facilities and humane treatment", "4. Valence": "Sympathetic."}

At this point, we grouped each set of 4 codes for each of the 400 clusters and generated a summary for each:

```
'Read a list of four themes from a cluster of passages ### ' + str(codes)+ ' ### Step by step, answer
the following: 1 Are all of these themes both present and relevant to attitudes towards refugees in
Malaysia? "All are."/"None are."/"Some are.". If irrelevant, return none to all further questions.  2.
If relevant, return the overarching theme as it relates to attitudes towards refugees in Malaysia, or
return None. Do not give a generic theme like "attitudes towards refugees in Malaysia", but provide a
specific and detailed theme. If relevant, 3. Whose attitudes are being reflected? Examples: the
government, Malaysians, NGOs, the author. 4. Who is the target of the attitudes? Examples: the
Rohingya, the government, UNHCR. 5. What is the overall valence, if any? Finally, Respond ONLY in the
following python dictionary format: {"1. Are Passages Relevant?":"All are."/"None are."/"Some are.","2.
Theme": stringval1/None, "3. Whose Attitude?":stringval2,"4. Target":stringval3,"5. Valence":
"Sympathetic."/"Hostile."/"N/A"}'
```

Here is the summary for the four codes given earlier:

{'1. Are Passages Relevant?': 'All are.', '2. Theme': 'Attitudes towards repatriation, citizenship, and humane treatment of Rohingya refugees in Malaysia', '3. Whose Attitude?': 'The speaker, Saifuddin, Rohingyas', '4. Target': 'Myanmar and Bangladesh, Rohingya refugees, Return home with facilities and humane treatment, Burmese government', '5. Valence': 'Sympathetic.'}

This leads us to our final prompt, which is run for the entire corpus:

```
'Read this passage from a news article ### ' + str(excerpt) + ' ###  If relevant, give the theme of
this SPECIFIC passage as it embodies, relates to, or reflects attitudes towards refugees in Malaysia.
The following summary of the excerpted article may provide context for the passage (e.g. who is being
discussed and where events are occurring): ### ' + str(summary) + ' ### Here is an overview of how
several passages similar to this one have been coded: ### ' + str(relevant) + ' ### DO NOT copy this
coding verbatim, but use it as reference and be careful if only some or none of the similar passages
were deemed relevant. Before answering, analyze step by step: 1. in 12 words or less, return the theme
(if relevant) as it relates to attitudes towards refugees in Malaysia, or return None. Do not give a
generic theme like "attitudes towards refugees in Malaysia", but provide a specific single theme. If
irrelevant, return None for all further questions. If relevant, 2. Whose attitudes are being reflected?
Examples: the government, Malaysians, NGOs, the author. 3. Who is the target of the attitudes?
Examples: the Rohingya, the government, UNHCR. 4. What is the valence of attitudes towards the target,
if any?:  "Sympathetic.", "Hostile.", or "N/A". Once again, the passage to code is ### ' +
str(excerpt)+ ' ###  Finally, Respond ONLY in the following python dictionary format: {"1. Theme":
None/stringval1, "2. Whose Attitude?":None/stringval2,"3. Target":None/stringval3, 4. Valence":
"Sympathetic."/"Hostile."/"N/A"}'
```

Here is an example of the output of this prompt, using the same passage as shown in AI Coding I:

> *{"1. Theme": "Highlighting past efforts to assimilate refugees", "2. Whose Attitude?": "Malaysia, collectively with many other nations", "3. Target": "Bosnian refugees", "4. Valence": "Sympathetic."}*

### *Coding Review II*

We then asked the reviewers to assess the new codes, while keeping the sample of passages to review fixed. Reproduced below is the prompt provided to the reviewers:

> *Just a reminder of the AI coding task:*
>
> *RQ:*
> *study the attitude towards refugees in Malaysia in Malaysian news media OR*
> *does this article have attidue [sic] towards refugees*
>
> *Same articles were used but the excerpts are of different lengths, compared with the previous version. Any excerpts below 100 or 150 characters are excluded. -- Clarification here: they were dropped because, in response to the problems with irrelevant passages in the first pass, we dropped from[sic] passages from consideration. Those include clearly irrelevant topics like subscription requests, as well as all passages less than 100 characters long.*
>
> *Now, the coding got 4 categories:*
> *1. Theme*
> *2. Whose attitudes towards refugees*
> *3. Who is the target of that attitude*
> *4. Valence: Hostile/Sympathetic*
> *The second to fourth categories is to help guide to assess whether the theme is accurate*
>
> *You will find some excerpts are not coded, this is*

*because we already highlighted it as irrelevant last time*

*Column*
*1. Assessment (Strongly Disagree, Disagree, Unsure, Agree, Strongly Agree) - Assess how much you agree*
*2. Relevant (Yes = 1, No = 0) - should this excerpt be coded? Is this excerpt relevant to the RQ?*
*3. Unrelated Code (Yes = 1, No = 0) - Is this code unrelated to the RQ?*
*4. Comments - whatever thoughts you have*

Clarifications were provided in communications documented in Appendix A.

At the conclusion of the reviewers' independent assessment of ChatGPT's coding, we convened a second round of discussions with all the reviewers. We compared and contrasted the reviewers' assessments to: 1) identify passages where the reviewers unanimously agreed (rated as either 'Agree' or 'Strongly Agree') to the coding, 2) identify passages where the reviewers unanimously disagreed (rated as either 'Disagree' or 'Strongly Disagree' to the coding, and 3) highlight passages where at least one reviewer assessed the coding differently from the other reviewers. In our discussion with the reviewers, we reviewed the codes that they unanimously disagreed to, to understand the reasons why and subsequently identify possible actions to further refine the instructions given to ChatGPT. We also reviewed the codes where there were disagreements between the reviewers and asked each reviewer to explain how they arrived at their assessment for the code. During this discussion, each reviewer described their interpretation of the code, the basis by which they provided an assessment to the code, and any uncertainties that they had in reviewing the code and its attached passage. The reviewers also discussed their expectations of ChatGPT's ability to code, their definitions of key terms and phrases, and other concerns that they had with the coding. The discussion was also supported by diaries that were updated by the reviewers throughout the assessment process. We moderated this discussion and resolved all disagreements in the assessments of the codes.

Having achieved a consensus in the assessments of the coding, we convened a further discussion in which the reviewers discussed their assessments of passages GPT coded as either relevant or irrelevant. Subsequently, we asked them to complete a final task in two steps: first, assess whether each "Irrelevant" passage was obviously irrelevant without reference to the article it came from, and then review the passages at least one reviewer deemed not obviously irrelevant in the context of the excerpted article. The instructions for the first step are reproduced below:

*Hi all, we've decided that we need another round of review! 🎉🎉🎉 Specifically, whether passages that were coded as irrelevant are actually irrelevant. First, we'd like you to choose*

*whether each of the 153 passages coded "None" are obviously irrelevant to the research question, namely: attitudes towards refugees in Malaysia.*

*This time, we want you to assess this ONLY on the basis of the passage text. If it's clear that, no matter the context provided by the article, this passage should not be coded, choose True in the "Obvious that this text should not be coded (T/F)" column*

*If it is clearly relevant or might be relevant depending on the context (e.g. there is a pronoun and the identity of that person might make it relevant), choose False*

And for the 9 passages that were marked as possibly relevant by at least one reviewer, the instructions were as follows:

*Can each of you go through these 9 passages, and, now in the context of the summary and full article, assess whether each passage is itself relevant to attitudes towards refugees?*

*Answer this in terms of strongly agree to strongly disagree.*

## Results

While the methodology is somewhat complex, the results can be related quite simply. For the first round of coding, there was a miscommunication regarding the assessment of passages GPT deemed "Irrelevant", leaving these review results unusable, but we present below the **consensus rate**, which we define as the percentage of passages where at least two out of three reviewers agreed or strongly agreed with GPT's coding, for the subset of the sample where GPT deemed the passage relevant. This is the **precision**[7] of the model.

| Results | Consensus Rate | 95% Confidence Interval (Clopper-Pearson) |
|---|---|---|
| Precision | 0.4961 | (0.4468, 0.5455) |

*Table 2: First round coding model performance.*

The result here is unsatisfactory. Less than half of the codes provided are deemed reasonable.

We also calculate inter-rater reliability (IRR) via Kendall's Tau-b, again only for the passages deemed relevant. The results suggest very limited consensus at the first stage.

---

[7] Formally, precision=True Positives/(True Positives + False Positives), where we treat the reviewer consensus as the ground truth.

| Results | Kendall's Tau-b (GPT coded relevant) | p-value |
|---|---|---|
| Raters 1&2 | 0.180 | 0.000 |
| Raters 2&3 | 0.383 | 0.000 |
| Raters 1&3 | 0.275 | 0.000 |

*Table 3: First coding inter-rater reliability*

For the final round of coding, we provide five key statistics for the consensus rate: **accuracy**[8], which is the consensus rate for the entire sample regardless of GPT's coding decision, precision, the consensus rate for the portion of the sample where GPT provided a code; **recall**[9], the fraction of passages where there was a consensus for relevance and GPT also coded the passage relevant; $F_1$ **score**[10], the harmonic mean of precision and recall; and **negative predictive value**[11], the consensus rate for the portion of the sample where GPT deemed the passage irrelevant.

| Results | Value | 95% Confidence Interval (Clopper-Pearson) |
|---|---|---|
| Accuracy | 0.8116 | (0.7729, 0.8478) |
| $F_1$ score | 0.8248 | (0.7854, 0.8622) |

---

[8] Accuracy = (True Positives + True Negatives)/(True Positives + True Negatives + False Negatives + False Positives), where we treat the reviewer consensus as the ground truth.
[9] Recall=True Positives/(True Positives + False Negatives), where we treat the reviewer consensus as the ground truth.
[10] $F_1$ Score=2/((1/Recall)+(1/Precision)).
[11] Negative predictive value=True Negatives/(True Negatives + False Negatives), where we treat the reviewer consensus as the ground truth.

| | | |
|---|---|---|
| Precision | 0.7192 | (0.6654, 0.7731) |
| Recall | 0.9738 | (0.9476, 0.9948) |
| Negative predictive value | 0.9675 | (0.9351, 0.9935) |

*Table 4: Final coding model performance*

The $F_1$ score is a commonly used metric of a classification model's predictive power, and the score of 82% suggests this methodology produces output (a code or an "irrelevant" classification) that human reviewers find reasonable a large majority of the time. The precision value shows us where the model falls short: about 30% of codes are not supported by a consensus. Recall and Negative predictive value show us where the model excels: identifying irrelevant passages. There are very few false negatives, leading to almost no errors in these measures. This should not be surprising: to succeed in generating a true negative, our model need only correctly assess relevance. To succeed in generating a true positive, the model needs to both assess relevance and then provide a specific code, a much more difficult task. Appendix A provides additional analysis of model performance.

We again measure IRR via Kendall's Tau-b. Here we see a dramatic increase in consensus relative to the first round. One major reason is that we include the entire dataset, not just the passages deemed relevant by GPT. Unsurprisingly, just as assigning a good code is a more difficult task than identifying an irrelevant passage, agreeing on what constitutes a good code is more difficult than agreeing a passage is irrelevant. Thus, consensus is almost universal for passages deemed irrelevant, but lower for coded passages. Even so, we still see a marked increase in consensus among coded passages relative to round one. One likely reason for this was that reviewers explicitly rerated each code based on the consensus building discussion in the final round. This suggests that the precision measured for the first version of the model may be less credible than that measured for the final version of the model.

| Results | Kendall's Tau-b (All coding) | Kendall's Tau-b (GPT coded relevant) | p-value (rounded value identical for both cases) |
|---|---|---|---|
| Raters 1&2 | 0.950 | 0.917 | 0.000 |
| Raters 2&3 | 0.780 | 0.519 | 0.000 |
| Raters 1&3 | 0.763 | 0.473 | 0.000 |

*Table 5: Final coding inter-rater reliability*

Discussion

In this novel method of AI coding that we propose, it is important to note the role and contribution of the (human) reviewers in the process. While we initially saw the role of the reviewers as simply being objective assessors and validators of the coding done by the AI, we quickly realized that the discussions between us and the reviewers led to substantial changes in the AI coding process and improvements in the coding performance. In effect, the coding performed by ChatGPT in our method is substantially human-informed. The inclusion of the reviewers' feedback on ChatGPT's coding, as well as the insights from the multiple discussions to resolve disagreements between the reviewers on how well ChatGPT coded the passages provided to it allowed us to have an iterative approach to coding that resembles the process that typically takes place between multiple human coders working on a coding framework where ongoing discussions lead to revisions, reinterpretations, and redefinitions of codes and terms until a consensus is reached and a final coding framework can be applied to the rest of the data. To our great interest, we also observed that in the discussions to resolve disagreements in coding assessment, one or more of the reviewers would often 'defend' the coding performed by ChatGPT, effectively giving ChatGPT a 'voice' in the discussions with the other (human) reviewers. It is important to note, however, that using this 'human-informed' approach to AI coding requires considerable work with the (human) reviewers, in addition to the work necessary to prepare the AI for coding. We noted from the reviewers' diaries, that throughout the process, the reviewers struggled with the feeling of uncertainty when assessing the coding done by ChatGPT as well as the discomfort of having to articulate and defend their method of rating in order to achieve consensus with the other reviewers. The reviewers also struggled with the back-and-forth discussions in order to reach a common understanding of key terms and to standardize a common approach to rating among themselves. This suggests to us that the reviewers have to be prepared early on to trust the iterative process, and the expectation that they will have to perform the task with some level of uncertainty before a final approach is developed.

# Appendix

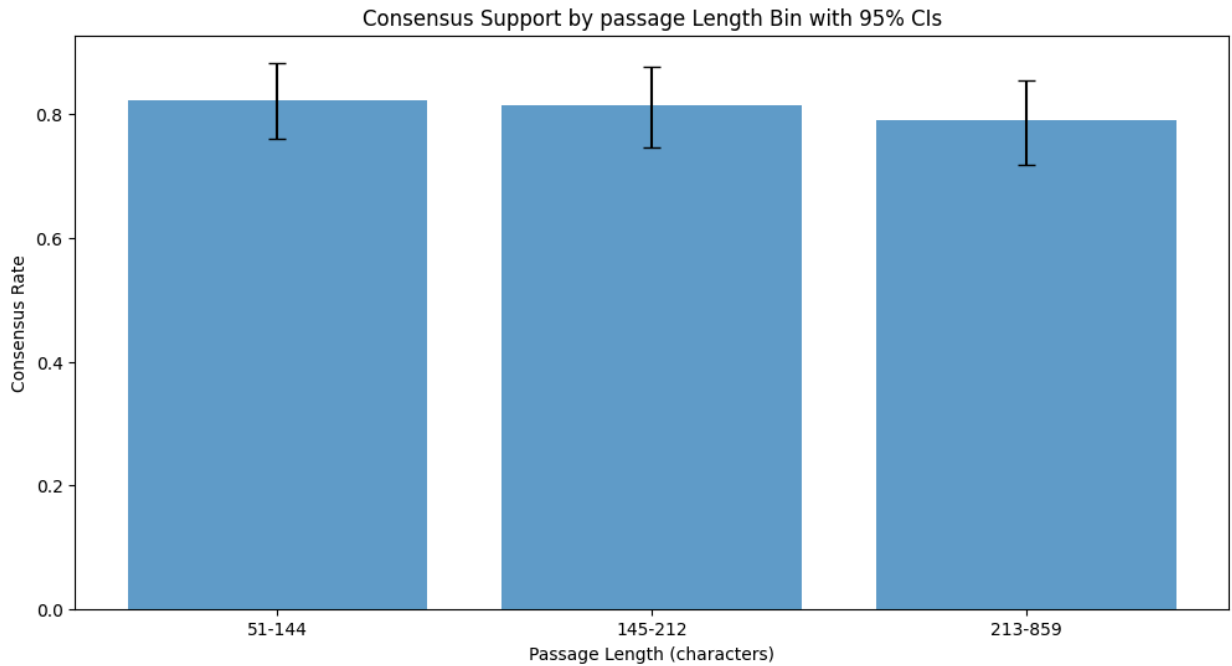

Figure A1: Consensus Support by document Length Bin with 95% CIs

Figure A1 breaks down the second round consensus rate by passage length. One might expect that shorter passages would perform worse due to a lack of context, or longer passages due to the potential for multiple codes within them or attention issues in the LLM with such a large document. However, we find no noticeable differences despite a wide range of passage lengths.

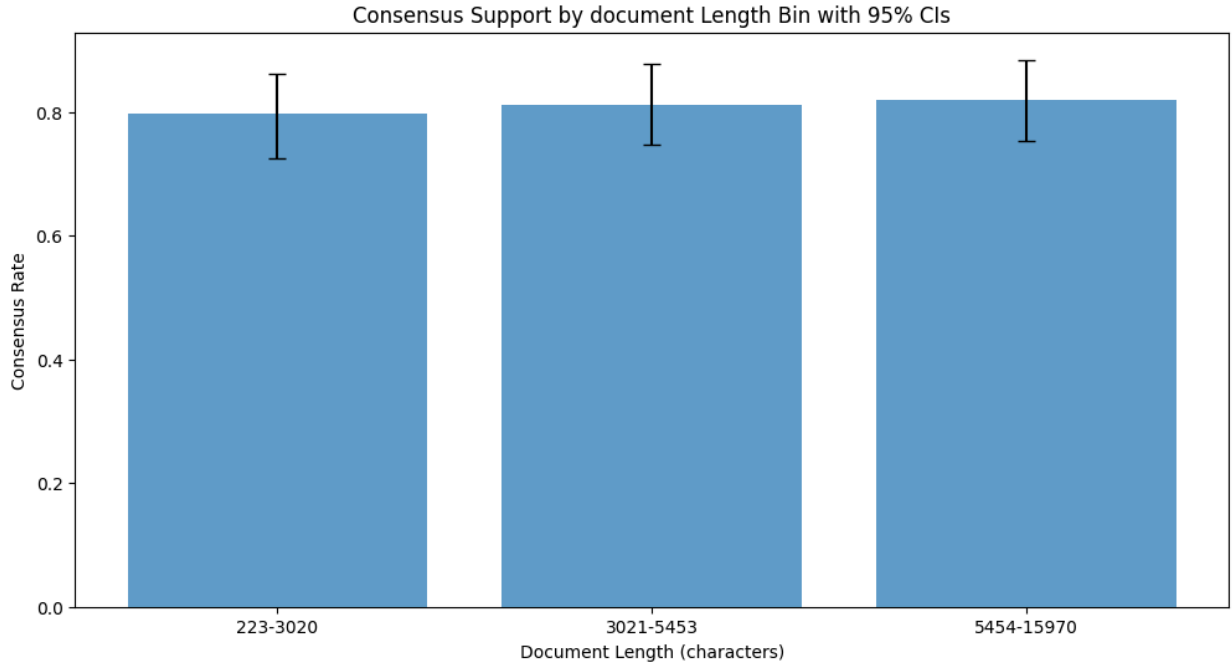

Figure A2: Fraction of Support > 1 for Each Summary Length Bin with 95% CIs

Figure A1 breaks down the second round consensus rate by document length. One might expect that longer documents would be harder to code, as there would be more context outside of the passage and the short summary would be less able to capture that extra context. However, we find no noticeable differences despite a wide range of document lengths.

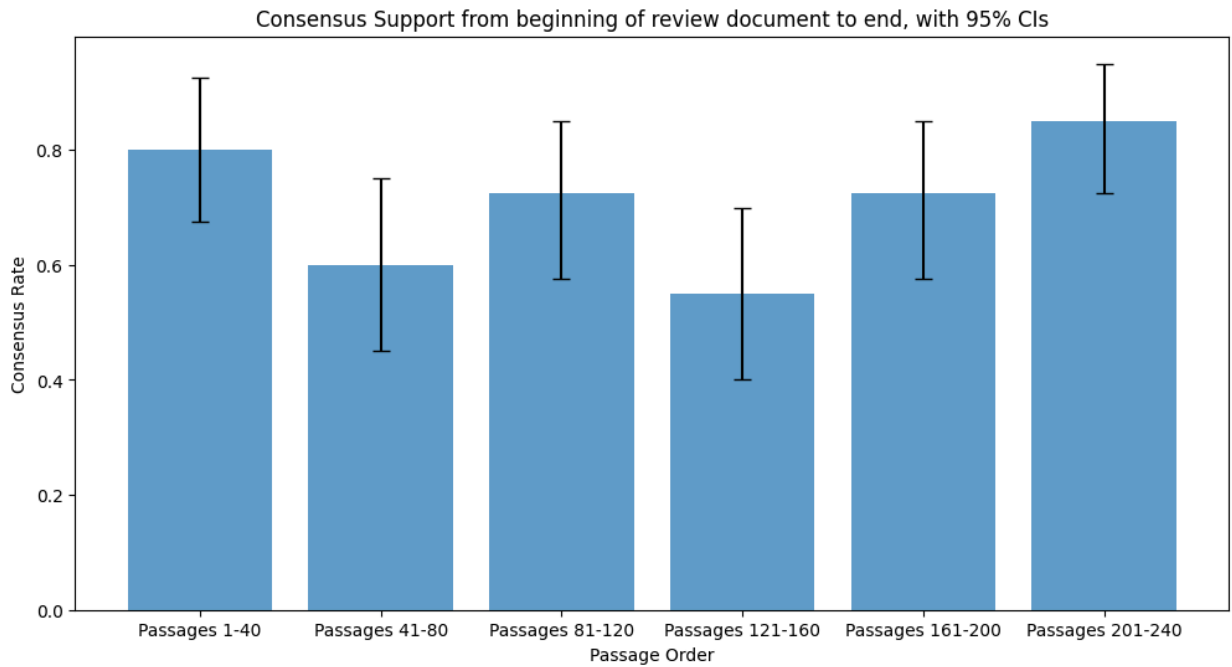

Figure A3: Consensus Support from beginning of review document to end, with 95% CIs

Figure A1 breaks down the second round consensus rate across the reviewing document. Reviewers were all given the same document with coded passages presented in the same order. Assuming they worked top to bottom, reviewers might be expected to change their assessments over the course of this process. While we see some variability across bins, there does not seem to be a clear trend. This may be due to the fact that the reviewers had already completed the process once, and thus their reviewing process had stabilized. It is also important to note that, to facilitate easier coding, passages were grouped by cluster in the review document. Therefore the order of the passages was nonrandom and systematic variation in consensus rate would not necessarily be evidence of changing reviewer standards.